\documentclass[runningheads]{llncs}
\usepackage[T1]{fontenc}
\usepackage{graphicx}
\usepackage{booktabs}
\usepackage{amsfonts}
\usepackage{color}
\usepackage{amsmath}
\usepackage{biblatex}
\newcommand{\todo}[1]{}
\renewcommand{\todo}[1]{{\color{red} TODO: {#1}}}
\newcommand{\olda}[1]{}
\renewcommand{\olda}[1]{{\color{blue} {#1}}}
\newcommand{\ondrej}[1]{}
\renewcommand{\ondrej}[1]{{\color{magenta} {#1}}}

\usepackage[breaklinks=true,colorlinks,bookmarks=false]{hyperref}
\begin{document}

\title{Leveraging Point Transformers for Detecting Anatomical Landmarks in Digital Dentistry}
\titlerunning{Transformers for Detecting Anatomical Landmarks in Digital Dentistry}
%
\author{Tibor Kubík\inst{1,2}
\and
Oldřich Kodym\inst{1}
\and
Petr Šilling\inst{1,2}
\and
Kateřina Trávníčková\inst{1}
\and
Tomáš Mojžiš\inst{1}
\and
Jan Matula\inst{1}
}
\authorrunning{T. Kubík et al.}
%
\institute{
TESCAN 3DIM, s.r.o., Brno, Czech Republic\\
\email{\{tibor.kubik,oldrich.kodym,petr.silling,katerina.travnickova, tomas.mojzis,jan.matula\}@tescan.com}\\
\url{https://tescan3dim.com/} \and
Department of Computer Graphics and Multimedia, Brno University of Technology, Brno, Czech Republic\\
\email{\{ikubik,isilling\}@fit.vut.cz}}
\maketitle              
\begin{abstract}
The increasing availability of intraoral scanning devices has heightened their importance in modern clinical orthodontics. Clinicians utilize advanced Computer-Aided Design techniques to create patient-specific treatment plans that include laboriously identifying crucial landmarks such as cusps, mesial-distal locations, facial axis points, and tooth-gingiva boundaries. Detecting such landmarks automatically presents challenges, including limited dataset sizes, significant anatomical variability among subjects, and the geometric nature of the data.
We present our experiments from the 3DTeethLand Grand Challenge at MICCAI 2024. Our method leverages recent advancements in point cloud learning through transformer architectures. We designed a Point Transformer~v3 inspired module to capture meaningful geometric and anatomical features, which are processed by a lightweight decoder to predict per-point distances, further processed by graph-based non-minima suppression. We report promising results and discuss insights on learned feature interpretability.
\end{abstract}
\section{Introduction}
\label{sec:introduction}
Anatomical landmark detection has been, for years, a crucial step in many 2D and 3D medical imaging tasks~\cite{cbctteeth-landmarks,handxray-landmarks,cephalo-landmarks,echocardiograms-landmarks}.
The growing availability of intraoral scanning devices has made automatic landmark detection increasingly important in modern clinical orthodontics, where clinicians employ advanced Computer-Aided Design techniques and 3D models to develop patient-specific treatment plans.
There, crucial landmarks include features such as cusps and mesial-distal locations, facial axis points, and tooth-gingiva limits, which are important for determining the alignment and positioning of the tooth in the dental arch, upper and lower teeth's occlusion relationship, orientation, angulation, and inclination of teeth.
Landmark detection in 3D orthodontic scans is difficult for the very same reasons as any other task from 2D medical imaging or (CB)CT/MRI scan analysis: limited dataset sizes and significant anatomical variability across subjects.
The fact that the optical impressions are represented in the form of 3D shapes adds a new layer of complexity as conventional operations in the non-Euclidean domain are not well-defined~\cite{bronstein2021geometricdeeplearninggrids}.

To advance the field of 3D medical shape learning, in this paper, we present our experiments conducted as part of our participation in the 3DTeethLand~\cite{ben_hamadou_2024_10991302} Grand Challenge at MICCAI 2024.

Our method is primarily based on recent advancements in point cloud learning leveraging transformer architectures. A \emph{PTv3}~(Point Transfomer v3)~\cite{wu2024pointtransformerv3simpler} module is designed to first capture meaningful geometric and anatomical features, which are then transferred to a compact MLP-based block trained to regress per-point distances represented as distance maps. Final point positions are extracted using a custom topology-driven graph-based non-minima suppression algorithm calibrated to individual landmark classes.
Our method is robust to rigid transformations due to data augmentation and is inherently resilient to minor morphological variations, as the samples underwent geometric morphing during training.

The presented work achieves promising results. It achieves average precision of \(0.64\) and average recall of \(0.64\) for error thresholds in interval between \(0\)~and \(2\) mm.
Apart from the main contributions, we also present further insights on the interpretability of the learned features from the encoder and the influence of the form of the inferred distance maps.

To facilitate further research in this area, we made our code publicly available on GitHub.\footnote{\href{https://github.com/tescan-3dim/PTv3-for-detecting-anatomical-landmarks-in-dentistry}{https://github.com/tescan-3dim/PTv3-for-detecting-anatomical-landmarks-in-den\-tistry}}

\subsection{Related Works}
\label{sec:related-works}
\subsubsection{3D shape learning.} One approach to classifying 3D deep learning is by the type of representation employed during the learning process. \emph{Projection-based} methods project 3D geometry onto various image planes and utilize well-established 2D CNN backbones and techniques to extract features~\cite{8100174,le:mvrnn,su2015multiviewconvolutionalneuralnetworks}. \emph{Voxel-based} approaches voxelize input shapes into 3D grids and apply 3D convolution and pooling operations, or its sparse versions, on such grids~\cite{chen2023voxelnext,7353481}.
Numerous efforts have emerged in the literature, involving either direct extension of basic deep learning operations on mesh structure~\cite{10.1145/3306346.3322959,10.1145/3506694,liang2022meshmaemaskedautoencoders3d} or more \textit{unorthodox} representations~\cite{dong2023laplacian2mesh,9713880,sharp2022diffusionnetdiscretizationagnosticlearning}.
A significant body of work focuses on deep learning approaches over point clouds~(\emph{point-based}), primarily driven by the fact that this format is the output of raw scanning of real-world objects.
Initial research in this area encounters challenges similar to that of other representations: the necessity for substantial undersampling due to high computational costs or the need to capture both local and global contexts~\cite{li2018pointcnnconvolutionmathcalxtransformedpoints,qi:pointnet,qi:pointnet++,wu2020pointconvdeepconvolutionalnetworks}.
With the shift towards transformer-based methods coupled with point serialization techniques, many works introduce memory-efficient frameworks with large receptive fields~\cite{10.1145/3592131,wu2022pointtransformerv2grouped,yu2022pointbertpretraining3dpoint,zhao2021pointtransformer}, with the latest advancement being \emph{Point Transformer v3}~(PTv3)~\cite{wu2024pointtransformerv3simpler}, the fastest and simplest state-of-the-art method in 3D vision tasks like indoor and outdoor semantic segmentation, indoor instance segmentation, and outdoor object detection.

\subsubsection{Automatic 3D digital dental landmarking.} 
There are numerous methods for landmark detection in 3D dental scans relying either on conventional approaches~\cite{WOODSEND2021104819} or deep learning techniques over various representations~\cite{Kubk2022RobustTD,10067081,9789163}. Wei et al.~\cite{WEI2022102077} trained a framework to detect landmarks and regress tooth axis. For the former, authors propose a multi-scale feature extraction module based on point cloud analysis that predicts per-point distance fields with K-Means clustering post-processing. The main limitation is the need to know the total number of landmarks beforehand to set the constant $K$, which is typically unknown in practice and should ideally be determined by the algorithm itself. 

In this work, we address key limitations of previous approaches to 3D dental landmarking by utilizing PTv3, a state-of-the-art point cloud backbone that enables landmark prediction without the need for a predefined number of landmarks. Furthermore, we introduce a novel topology-driven non-minima suppression technique, which is versatile, seamlessly integrable into any detection network, and overcomes the challenge of high false-positive rates seen in previous methods.

\subsection{Contributions}
\label{sec:contributions}
Our contributions to the field are as follows:
\begin{itemize}
    \item We evaluate one of the strongest and most recent backbones in point cloud learning, PTv3~\cite{wu2024pointtransformerv3simpler}, within the domain of digital dentistry and demonstrate its ability to learn meaningful features despite the constraints of a limited medical dataset.
    \item We present a topology-driven non-minima suppression post-processing technique that effectively reduces false-positive landmark predictions and can be easily integrated into other frameworks~\cite{WEI2022102077} to make them more applicable in real clinical scenarios.
\end{itemize}

\begin{figure}[htbp]
\centering
\includegraphics[width=\linewidth]{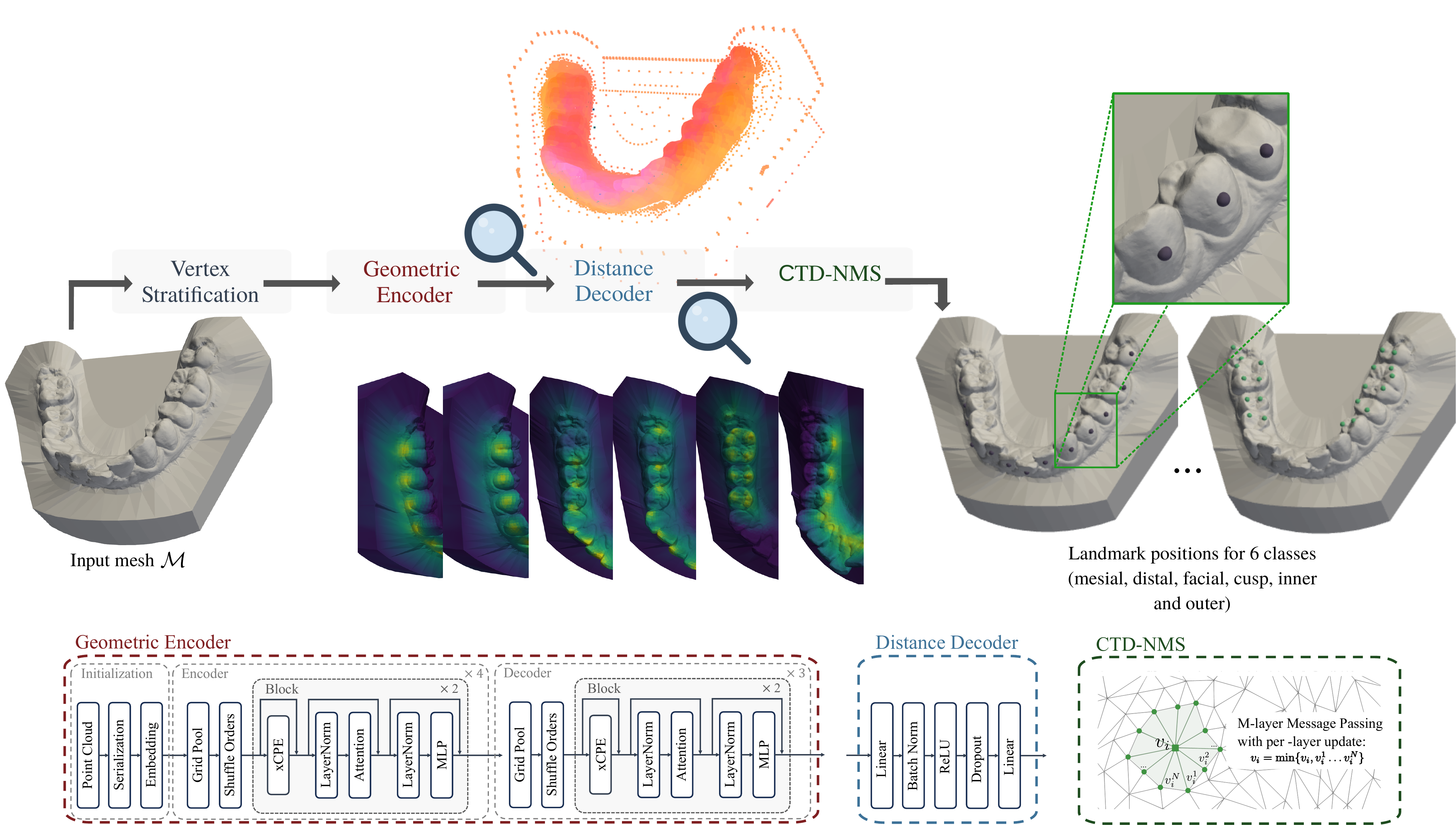}
\caption{Outline of the proposed method. The pipeline consists of vertex stratification followed by a geometric encoder-decoder architecture that generates dense distance maps (visualized as heat maps), and CTD-NMS (Calibrated Topology-Driven Non-Maximum Suppression) for final landmark positioning.}
\label{fig:method-outline}
\end{figure}

\section{Method}
\label{sec: method}
The outline of our method is depicted in Figure~\ref{fig:method-outline}.
Given a set of $N$ vertices, $\mathcal{V} \in \mathbb{R}^{N \times 3}$, a set of edges $\mathcal{E}\subseteq \mathcal{V}^2$, and a~set of faces $\mathcal{F} = \{(k, l, m) \mid k, l, m \in \mathcal{V} \land (k, l), (l, m), (m, k) \in \mathcal{E\}}$, the input mesh representing a single 3D dental scan is denoted as $\mathcal{M} = (\mathcal{V}, \mathcal{E}, \mathcal{F})$.
The method does not require color or texture information.
The objective of the method is to model a function that generates landmark positions based on $\mathcal{M}$ as points in $\mathbb{R}^3$.  These landmarks are categorized into six well-defined classes: \emph{mesial}, \emph{distal}, \emph{cusp}, \emph{inner}, \emph{outer}, and \emph{facial}. The number of landmarks in each class varies from patient to patient, as it is influenced by the number of present teeth and individual dental morphology (for instance, tooth 36 in FDI World Dental Federation notation may have 4~cusp points in one case and 5 in another). Please refer to Figure~\ref{fig:method-outline} for their visualization. 

Unlike many existing approaches, which operate on sparse subsets of vertices or require specific mesh preprocessing, our method is designed to work effectively with dense point clouds derived directly from the input mesh.
Note that the framework is versatile and applicable across different anatomical contexts within orthodontics since it is jaw-type agnostic. In addition, the framework does not put any requirements in terms of mesh manifoldness, or tesselation.

At a high level, the framework consists of three main components: the \emph{geometry encoder}, the \emph{distance decoder}, and the \emph{topology-driven non-minima suppression}. A description of individual components follows.

\subsubsection{Geometry encoder.}
For the encoder, mesh connectivity is not considered, and the input is formed based on the vertex set $\mathcal{V}$. Since the number of vertices varies among subjects and some meshes are unnecessarily fine, we stratify the vertex point cloud to a fixed number of $M$ points, resulting in point cloud $\mathcal{V}_s \in \mathbb{R}^{M \times 3}$. We fix $M$ to $64\,000$, an empirically set value, in all of our experiments. $M$ vertices are selected randomly from the original set $\mathcal{V}$.  
It is important to note that other point-sampling techniques, such as \emph{furthest point sampling} (FPS), can be used to extract point clouds from mesh structures. While applying FPS is a common practice, we observed that such sampling methods often generate points on irrelevant parts of the mesh in the dataset. Dental structures were processed using a closing algorithm specifically designed for such data, generating a structure that resembles a \textit{platform}~(visible in Figure~\ref{fig:method-outline}). The FPS algorithm tends to sample a significant portion of the data from these artificial structures. In contrast, our approach effectively mitigates this issue, as the added structures consist of a small number of faces, resulting in a point cloud predominantly composed of points from teeth and gingiva. We use this sub-sampling technique during training phase but we use the full point clouds during validation and testing. We observed that the model generalizes well to higher amounts of points.

$\mathcal{V}_s$ can be enriched by additional per-point $d$-dimensional features, resulting in a point cloud $\mathcal{V}_e \in \mathbb{R}^{M \times (3+d)}$. 
We fix $d = 3$ since vertex normal information is concatenated to each point.

The geometric encoder is a mapping $\mathcal{V}_e \mapsto \mathbb{R}^{M \times h}$, where the value of $h$ defines the dimensionality of learned per-point features embedded in latent space.
Internally, the design tightly follows the U-Net-inspired encoder-decoder PTv3 architecture~\cite{wu2024pointtransformerv3simpler}.
The PTv3 architecture has superior feature extraction capabilities that allow the framework to capture meaningful geometric patterns in jaw shapes, setting it apart from prior methods.

\subsubsection{Distance decoder.} 
The decoder maps learned per-point features from the encoder to 6-channel per-point distances in $\mathbb{R}^{M \times 6}$. This allows the framework to detect landmarks within 6 different classes without requiring intra-class label assignment. 
In other words, the decoder assigns six scalars to each point in $\mathcal{V}_s$, defining the distance to the closest landmark in each category.
We wanted to avoid the direct regression of resulting point coordinates since their number is not fixed, and a special module is dedicated to point extraction.

This part of the network is modelled as a compact multi-layer perceptron (MLP) module. 
Based on empirical findings, the geometric encoder learns meaningful complex feature representations. Processing these features with a more \textit{heavyweight} decoder caused overfitting.

\subsubsection{Topology-driven non-minima suppression.}
Finally, the peaks representing the landmarks are extracted from the distance decoder for each class. A straightforward approach would be to extract the maxima positions from the predicted and thresholded distance maps. However, this method is not sufficiently robust, as it tends to produce clusters of false positive predictions. This issue arises due to the blurred output signal from the distance decoder, particularly when the input geometry itself provides limited information.

To tackle this issue, we define a \emph{calibrated topology-driven non-minima suppression} (\emph{CTD-NMS}) module that converts distance maps to cluster-free final landmark positions.
Unlike traditional post-processing approaches, which rely solely on local maxima detection and are prone to false positives, the CTD-NMS module directly incorporates the mesh topology for improved landmark localization.
CTD-NMS leverages the full geometry potential of $\mathcal{M}$ by operating on the mesh defined by its topology. 
The module is implemented as a graph-convolutional operator that, for each vertex $v_i \in \mathcal{V}$, updates its value with $\min\{v_i, v_i^1 \dots v_i^N \}$, where $\{v_i^1 \dots v_i^N \}$ is the set of vertices connected to $v_i$ in the mesh graph. After applying this operator to the graph iteratively for $K$ steps, we classify a vertex as a landmark vertex if it satisfies the following two conditions: first, the vertex value did not change after applying the non-minima suppression operator, and second, the distance value of the vertex is less than a threshold value $T$. Both $K$ and $T$ are hyperparameters dependent on the model quality, definition of distance map and typical mesh resolution in the dataset. We calibrate these hyperparameters using validation data, evaluating the model on all combinations of $K$ and $T$, where $K \in \{ 6, 10, 13, 17, 21, 25, 28, 32 \}$ and $T \in \{0.1, 0.2,\dots, 0.8\}$. During the evaluation, we measure precision and recall and choose the combination of $K$ and $T$ with the highest product of precision and recall as the most optimal. Each landmark class has its own hyperparameter values, which are optimized together with the hyperparameters of other landmark classes.

\subsection{Data Augmentation}
\label{sec:preprocessing}
To improve the generalization ability, we increase the dataset size by augmenting its original samples. We randomly apply a rigid transformation matrix $A_{rig} = [R|t]$ on input meshes, where $R \in SO(3)$ denotes the rotation matrix and $t \in \mathbb{R}^3$ denotes the translation vector.
We rotate input samples by a random angle uniformly sampled within $[-0.5, 0.5]$ radians around an arbitrary axis. The vector $t$ displaces the mesh by translating it by a value uniformly sampled between $[-5, 5]$~mm.
Additionally, we apply the following non-rigid augmentations: (1) isotropic scaling of the shape by a scalar from range $[0.8, 1.2]$, and (2) geometrical morphing~\cite{TezzeleDemoMolaRozza2020PyGeM} using \emph{Free-Form Deformation} (FFD) technique~\cite{10.1145/15922.15903}. In this process, we utilize bounding boxes of the meshes as deformation grids, applying control point displacements within a range of $[-5, 5]$ mm. For each original sample, two additional deformed versions are generated using this approach. All augmentations are consistently applied to the corresponding ground truth landmark positions, ensuring alignment with the modified mesh geometries.

\subsection{Preparation of Distance Map Labels}
\label{sec:preprocessing}
Let $\mathcal{V}$ be the set of vertices of $\mathcal{M}$ and $\mathcal{P}_c = \{ p_j \}_{j=1}^{M}$ the set of ground truth points for a given landmark class $c$. For each vertex $v_i \in \mathcal{V}$, the geodesic distance to the nearest point in $\mathcal{P}_c$ is defined as $d_i^c = \min_{p_j \in \mathcal{P}_c} \text{Geo}(v_i, p_j),$
where $\text{Geo}(v_i, p_j)$ denotes the geodesic distance~\cite{crane2020surveyalgorithmsgeodesicpaths} between vertex $v_i$ and point $p_j$.

To mitigate the influence of distant vertices, the distance attribute $d_i^c$ is clamped to a maximum threshold value $\tau$: $\hat{d}_i^c = \min(d_i^c, \tau).$ We set $\tau = 15$\,mm in our setup.
This clamped distance value $\hat{d}_i^c$ is then used as a per-vertex attribute for each vertex in $\mathcal{V}$, ensuring that the network prioritizes learning relevant local geometric information.

Since such distance maps, when approximated by an estimator model, often tend to be oversmooth to the point where two adjacent distance peaks may be difficult to distinguish, it may be beneficial to further post-process the distance values to achieve maps that are sharper around the target landmarks. We achieve this by simply taking the square root of the distance values and normalizing the result.

\section{Experimental Setup}
\label{sec:experimental-setup}

\subsection{Dataset}
\label{sec:dataset}
We conduct all our experiments using the \emph{Teeth3DS}~\cite{ben2022teeth3ds} dataset, which was first introduced in \emph{3DTeethSeg'22: 3D Teeth Scan Segmentation and Labeling Challenge}~\cite{ben20233dteethseg}.
A total number of \textit{1800} 3D intra-oral scans of 900 patients in the form of triangle meshes has been collected by orthodontists and dental surgeons.
For the purpose of the 3DTeethLand MICCAI2024 challenge, the authors provided annotations of landmarks for 200 out of the total number of data samples. Thus, the remaining 1600 cases are not directly usable in purely supervised frameworks.
The cases are evenly distributed into maxillas and mandibles. All cases have been further anonymized, but the authors claim that the provided dataset follows a real-world patient sex and age distribution.

Since all provided cases are real orthodontic patients,  the overall statistics regarding missing teeth distribution are considered representative. 
It should be noted that this type of variability in the dataset is important since the number of landmarks that need to be detected varies based on the number of teeth in the arch. A high bias in the dataset could lead to many false positive detections in real clinical scenarios.

For more detailed information about the dataset, we direct the reader to the corresponding articles~\cite{ben2022teeth3ds,ben20233dteethseg}. 

\subsection{Evaluation Metrics}
\label{sec:evaluation-metrics}
A random subset of 20 validation samples was chosen for evaluating the model performance during experimentation. We compute the detection precision \(P\) at distance threshold \(t\) as \(P_t=\frac{TP}{TP+FP}\), where \(TP\) is the number of the detected landmarks that have a ground truth landmark within Euclidian distance \(t\), and \(FP\) is the number of detected landmarks that do not. Similarly, we compute the detection recall \(R\) at distance threshold \(t\) as \(R_t=\frac{TP}{TP+FN}\), where \(FN\) is the number of ground truth landmark that do not have a detected landmark within Euclidian distance \(t\). Finally, we report the averaged precision and recall over several distance thresholds for each landmark class separately. We sample distance thresholds from $[0, 2]$ linearly with step \(0.05\) mm.

\subsection{Training and Implementation Specifications}
\label{sec:implementation-specifications}
The network is optimized using batches of size $4$, AdamW optimizer~\cite{loshchilov2019decoupledweightdecayregularization}, a learning rate of $0.001$ and a weight decay of $0.001$.
Cosine annealing restarts are applied every $10\,000$ iterations. The model is trained by minimizing the Mean Squared Error (MSE) loss between the geodesic distance maps predicted by the distance decoder and the ground truth geodesic maps.

Our geometric encoder consists of a four-stage encoder and a three-stage decoder, with respective block depths $[2, 2, 2, 2]$ and $[2, 2, 2]$. The patch size is fixed to $1024$ in all stages.
The distance decoder comprises a single hidden layer with a dimensionality of $512$.
An increase in decoder capacity did not result in any increase in performance.

The experiments presented in this study were conducted on a machine equip\-ped with a 24GB NVIDIA GeForce RTX 4090 GPU and a 12-core AMD Ryzen 9 7900 CPU, paired with 64 GB of RAM. The operating system was Ubuntu 22.04 LTS. Python 3.10 was used in the experiments, and the primary libraries utilized include PyTorch, PyTorch Lightning, PyTorch Geometric, and Trimesh.

We trained the model for $100\,000$ iterations, which took approximately 15 hours.

\section{Results}
\label{sec:results}
We tested the aforementioned model configuration on regression of two different types of distance maps mentioned in Section \ref{sec:preprocessing}. We refer to the models trained for the respective tasks as the \textit{baseline} model for regression of the original clipped distance values, and the \textit{sharpened} model for regression of the square root-transformed distance values. The precision and recall metrics for the two models, described in Section \ref{sec:evaluation-metrics}, are shown in Table~\ref{tab:results}.
For completeness, we also report the calibrated values for the number of non-minima suppression operator steps and the calibrated detection thresholds in Table~\ref{tab:nms}. Note that due to the different distribution of distance values between the baseline and the sharpened models, the evaluated threshold values for the baseline model were higher than those presented in Section~\ref{sec: method}.

\begin{table}[]
\centering
\caption{Precision (P) and recall (R) values for individual landmark classes and the total precision and recall values averaged over mulptiple error thresholds.}

\begin{tabular}{l|ll|ll|ll|ll|ll|ll|ll} 
\toprule
          & \multicolumn{2}{l|}{Mesial}   & \multicolumn{2}{l|}{Distal}   & \multicolumn{2}{l|}{Cusp}     & \multicolumn{2}{l|}{Inner}    & \multicolumn{2}{l|}{Outer}    & \multicolumn{2}{l|}{Facial}   & \multicolumn{2}{l}{Total}      \\ 
\hline
          & P             & R             & P             & R             & P             & R             & P             & R             & P             & R             & P             & R             & P             & R              \\ 
\hline
Baseline  & 0.61          & 0.60          & 0.58          & 0.57          & \textbf{0.68} & \textbf{0.66} & 0.52          & 0.51          & 0.61          & 0.61          & 0.61          & 0.62          & 0.60          & 0.60           \\
Sharpened & \textbf{0.64} & \textbf{0.63} & \textbf{0.66} & \textbf{0.66} & 0.65          & 0.64          & \textbf{0.60} & \textbf{0.61} & \textbf{0.63} & \textbf{0.64} & \textbf{0.66} & \textbf{0.67} & \textbf{0.64} & \textbf{0.64}  \\
\bottomrule
\end{tabular}
\label{tab:results}
\end{table}

\begin{table}[]
\centering
\caption{The number of optimal non-minima suppression operator steps and the corresponding threshold values for each landmark class.}
\begin{tabular}{l|ll|ll|ll|ll|ll|ll} 
\toprule
          & \multicolumn{2}{l|}{Mesial} & \multicolumn{2}{l|}{Distal} & \multicolumn{2}{l|}{Cusp} & \multicolumn{2}{l|}{Inner} & \multicolumn{2}{l|}{Outer} & \multicolumn{2}{l}{Facial}  \\ 
\hline
          & St. & Th.                 & St. & Th.                 & St. & Th.               & St. & Th.                & St. & Th.                & St. & Th.                 \\ 
\hline
Baseline  & 13    & 2.59                & 21    & 2.17                & 10    & 1.76              & 10    & 2.17               & 10    & 1.76               & 10    & 2.17                \\
Sharpened & 13    & 0.70                & 25    & 0.70                & 10    & 0.40              & 10    & 0.40               & 13    & 0.50               & 10    & 0.40                \\
\bottomrule
\end{tabular}
\label{tab:nms}
\end{table}

\begin{figure}[htbp]
\centering
\includegraphics[width=\linewidth]{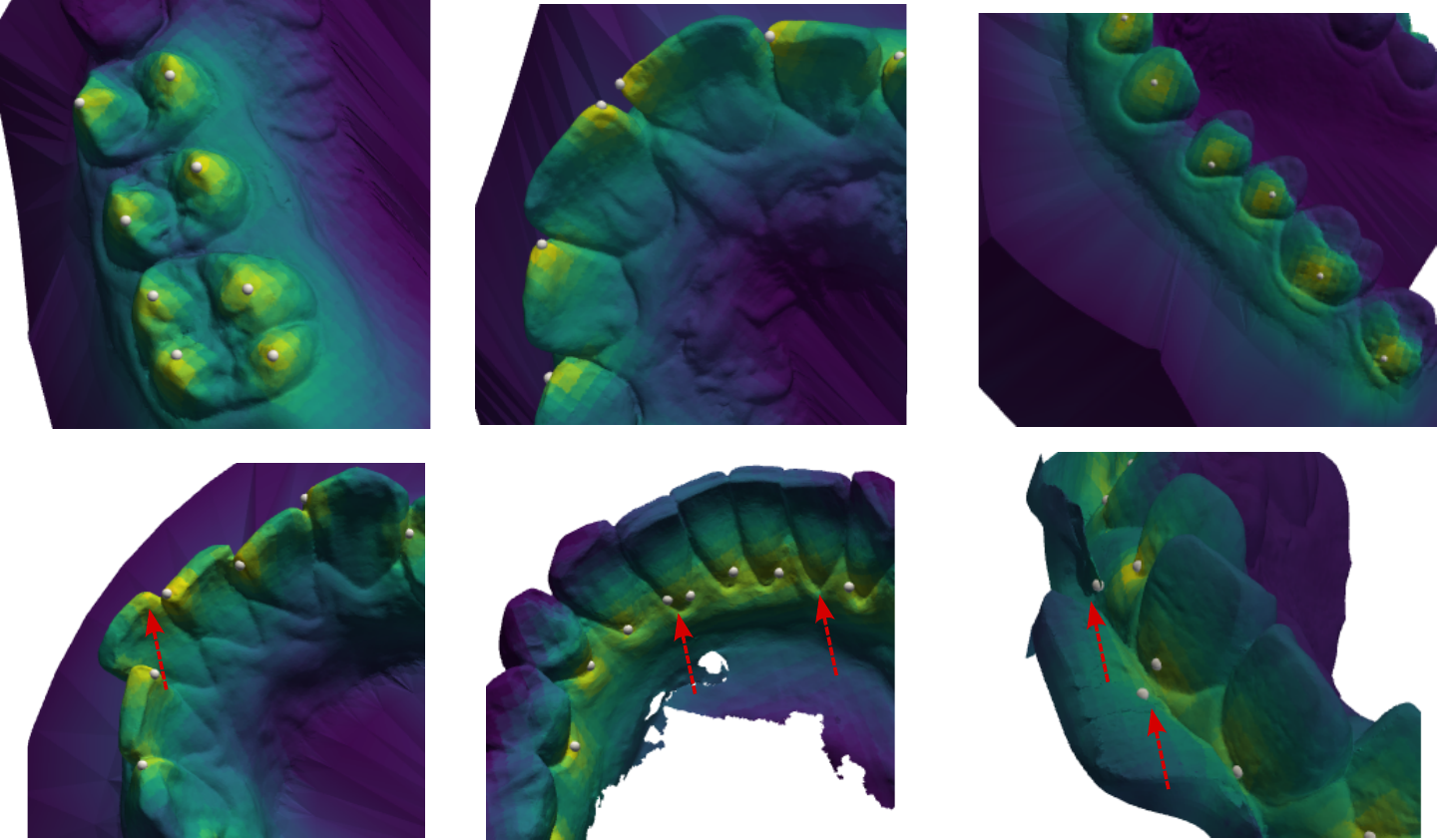}
\caption{Sample outputs on validation cases. Estimated distance maps, along with post-processed detections, are shown for a specific landmark class for each case. Top row: cases with satisfying results. Bottom row: failure cases.}
\label{fig:qualitative-results}
\end{figure}

\begin{figure}[htbp]
\centering
\includegraphics[width=\linewidth]{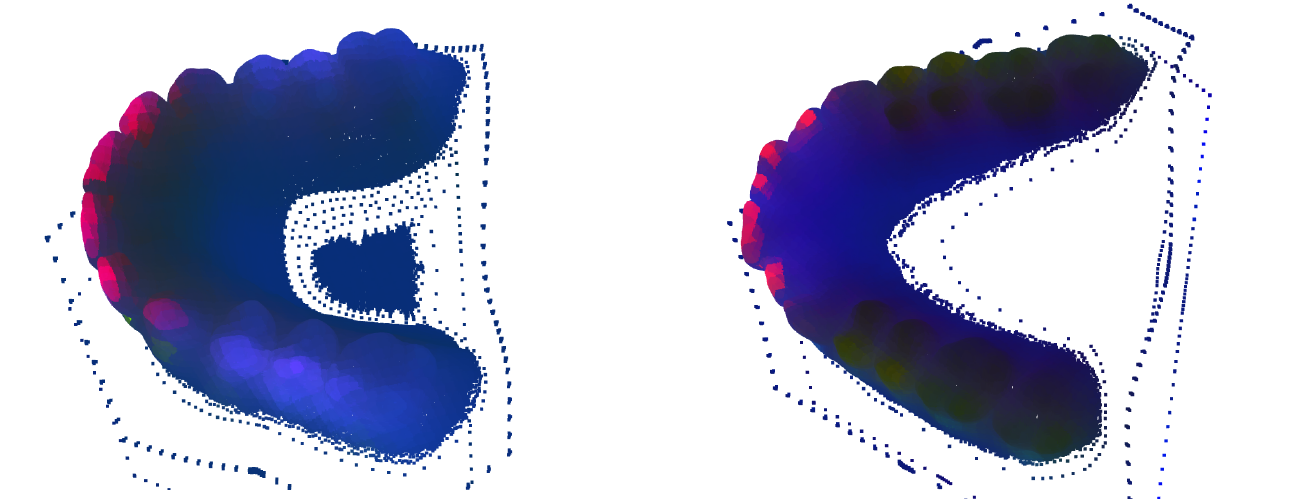}
\caption{Visualization of learned feature embeddings from the PTv3 encoder, projected onto RGB color space using PCA on subsampled vertices from two different validation cases.}
\label{fig:encoder-features}
\end{figure}

\subsection{Discussion}
\label{sec:discussion}
Sharpening the distance maps representing landmark positions on the jaw mesh brings significant increase in detection robustness for all landmark classes except for \emph{cusp}. Higher spatial gradients in near vicinity of the landmark positions force the neural network to focus its modeling capacity on the areas that are critical for correct post-processing outcomes.
Additional qualitative results are available in the supplementary material within the project repository.

Even though the results in most of the cases show high levels of accuracy and robustness (Figure \ref{fig:qualitative-results}, top row), there are still several types of errors in the predictions, such as imprecise inner point detections near crooked teeth or missing mesial point detections between frontal incisors, which are typically coupled very close together (Figure \ref{fig:qualitative-results}, bottom row). While the accuracy of the distal point detection is relatively high, the high calibrated value of non-minima suppression steps from Table~\ref{tab:nms} for this landmark class suggests that false detections need to be suppressed in the post-processing step.

In addition to the detection results, we inspected the geometric features computed by the geometry encoder before transforming them into distance values. As shown in Figure \ref{fig:encoder-features}, we projected the high-dimensional feature vectors of each point in the point cloud onto 3-dimensional RGB vectors using principal components analysis (PCA) for visualization purposes. Note that the geometry encoder implicitly learns to segment the buccal and lingual side of posterior teeth and recognize the incisal edges of frontal teeth.

The proposed model has 8.9 million trainable parameters and a compact size of 35.4 MB. Inference for a single intra-oral 3D scan, including the CDT-NMS post-processing step, takes 1.13 ± 0.01 seconds on a consumer-grade GPU (GeForce RTX 3080 Ti). This combination of fast inference and low memory footprint enables efficient real-time applications, such as surgical planning or diagnostics, within a clinical setting and without requiring specialized computing infrastructure.

\section{Conclusions}
\label{sec:conclusions}
In this paper, we presented the results of our participation in the 3DTeethLand Grand Challenge at MICCAI 2024. Our method, based on recent advances in transformer-based point cloud analysis coupled with topology-aware non-minima suppression, achieves promising results and extends the current state-of-the-art in the task of 3D dental landmark detection.
To our knowledge, this is the first attempt at using a point transformer-based architecture for the task of unconstrained landmark detection on 3D mesh surfaces.

\begin{credits}
\subsubsection{\ackname} We present our gratitude to our colleagues, including Ondrej, Braňo, David, and Michal, for their assistance and valuable insights in the preparation of this work. We also thank the company TESCAN 3DIM s.r.o. for providing the computational resources.
\end{credits}

\printbibliography

\end{document}